\pdfoutput=1 
\documentclass[10pt,twocolumn,letterpaper]{article}

\usepackage{wacv}
\usepackage{times}
\usepackage{epsfig}
\usepackage{graphicx}
\usepackage{amsmath}
\usepackage{amssymb}

\usepackage{algorithm}
\usepackage{algorithmicx}
\usepackage{algpseudocode}
\usepackage{url}
\usepackage{multirow}
\usepackage{array}
\usepackage{microtype}
\usepackage[usenames,dvipsnames]{color}
\usepackage{colortbl}
\usepackage{booktabs}
\usepackage[accsupp]{axessibility}  
\usepackage{authblk}

\definecolor{mycyan}{cmyk}{.1,0,0,0}
\definecolor{mypink}{rgb}{.99,.91,.95}

\usepackage[pagebackref=true,breaklinks=true,letterpaper=true,linkcolor=blue,citecolor=blue,colorlinks,bookmarks=false]{hyperref}

\def\wacvPaperID{867} 

\wacvfinalcopy 

\ifwacvfinal
\fi

\ifwacvfinal
\usepackage[breaklinks=true,bookmarks=false]{hyperref}
\else
\usepackage[pagebackref=true,breaklinks=true,colorlinks,bookmarks=false,citecolor=blue,linkcolor=blue]{hyperref}
\fi

\ifwacvfinal
\fi

\begin{document}

\title{Semi-supervised Multi-task Learning for Semantics and Depth}

\author{Yufeng Wang\textsuperscript{1}, Yi-Hsuan Tsai\textsuperscript{2}, Wei-Chih Hung\textsuperscript{3}, Wenrui Ding\textsuperscript{1},  Shuo Liu\textsuperscript{1}, Ming-Hsuan Yang\textsuperscript{4}  \\
\textsuperscript{1}Beihang University,
\textsuperscript{2}Phiar Technologies,
\textsuperscript{3}Waymo,
\textsuperscript{4}University of California at Merced\\
{\tt\small \{wyfeng, ding, liush\}@buaa.edu.cn, wasidennis@gmail.com, hungwayne@waymo.com, mhyang@ucmerced.edu }
}

\maketitle
\thispagestyle{empty}

\begin{abstract}
Multi-Task Learning (MTL) aims to enhance the model generalization by sharing representations between related tasks for better performance. Typical MTL methods are jointly trained with the complete multitude of ground-truths for all tasks simultaneously. 
However, one single dataset may not contain the annotations for each task of interest. 
To address this issue, we propose the Semi-supervised Multi-Task Learning (SemiMTL) method to leverage the available supervisory signals from different datasets, particularly for semantic segmentation and depth estimation tasks.
To this end, we design an adversarial learning scheme in our semi-supervised training by leveraging unlabeled data to optimize all the task branches simultaneously and accomplish all tasks across datasets with partial annotations.
We further present a domain-aware discriminator structure with various alignment formulations to mitigate the domain discrepancy issue among datasets. 
Finally, we demonstrate the effectiveness of the proposed method to learn across different datasets on challenging street view and remote sensing benchmarks.
\end{abstract}

\section{Introduction}
\label{sec:introduction}
Multi-Task Learning (MTL) aims to leverage information contained in multiple related tasks to improve the performance of each single task \cite{zhang2017survey}. The potential advantages of MTL over separate learning of each task can be attributed to the generalization ability by sharing representations among related tasks as well as the benefit of multiple sources with supervision. It has been widely used in numerous tasks in computer vision \cite{kokkinos2017ubernet}, 
natural language processing \cite{collobert2008unifiednlp}, and speech recognition \cite{michael2013mtl}, to name a few. 

Recently, deep convolutional neural networks (CNNs) have been successfully applied to dense prediction tasks such as semantic segmentation \cite{long2015fcn,chen2018deeplab} and depth estimation \cite{eigen2014depth,godard2017unsupervised}. 
One of the reasons for this success is the construction of large-scale and diverse datasets with pixel-wise annotations. 
Typical MTL methods train all tasks simultaneously within one dataset that contains the complete multitude of ground-truths.
However, in the real-world scenario, a single dataset usually does not contain all necessary ground-truths.
In addition, annotating the dataset for missing tasks entails significant effort and time, especially for the dense prediction tasks.
To tackle this issue, one can leverage different datasets that contain the corresponding annotations for each task.
Therefore, it is of great interest and importance to enable the network to leverage different supervision information from diverse datasets in the MTL framework \cite{kokkinos2017ubernet,li2018lwf}.

\begin{figure} [!t]
        \begin{center}
            \includegraphics[width=0.48\textwidth]{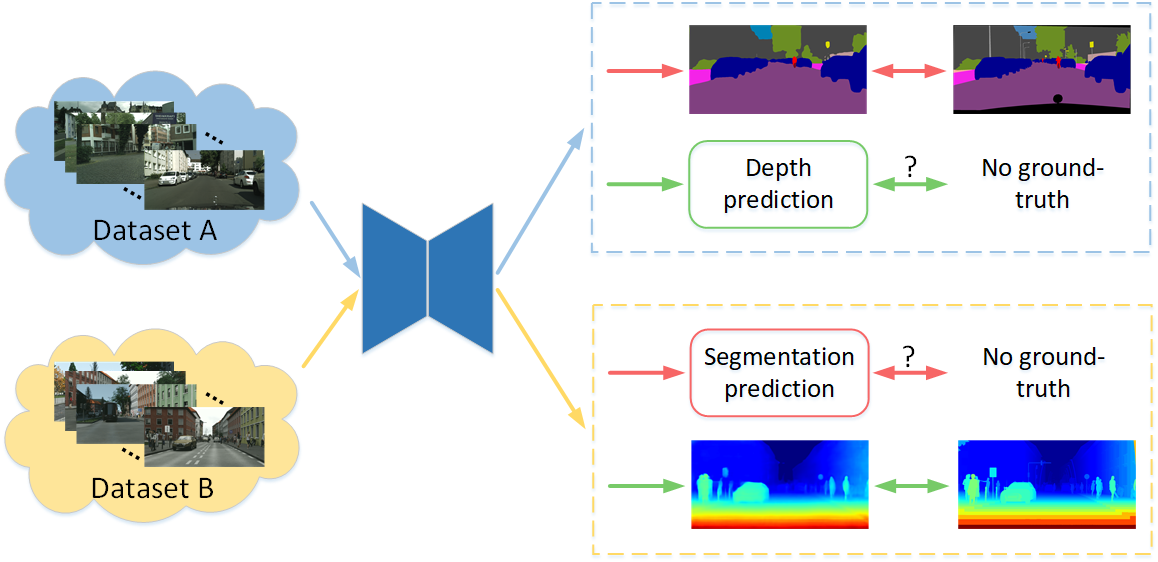}
    	\end{center}
    	\vspace{-3mm}
    	\caption{\textbf{Problem illustration.} Given two datasets $\mathcal{A}$ and $\mathcal{B}$, each of which only provides the annotation for partial tasks, we exploit the multi-tasking across datasets and mitigate the domain discrepancy to learn a more generalized model for all tasks on both datasets.}
    	\label{fig:motivation}
    	\vspace{-6mm}    	
    \end{figure}

We consider the setting where only disjoint datasets are at our disposal with partial ground-truths, $\emph{e.g.}$, dataset $\mathcal{A}$ developed for semantic segmentation and dataset $\mathcal{B}$ collected for depth estimation, as shown in Figure \ref{fig:motivation}.
%
%
It is in line with our intuition that a model can learn generalized representations from different datasets while ensuring at least one reliable supervision to train each task.
To this end, one straightforward approach is to learn from one dataset for one task at a time, and alternatively train the joint model for MTL \cite{kokkinos2017ubernet}.
Nevertheless, such a learning scheme does not consider domain gaps across datasets \cite{thai2018adaptseg,nath2018adadepth}.

In this work, we propose a Semi-supervised Multi-Task Learning (SemiMTL) method to expand the MTL setting for training the joint model across different datasets with partial annotations.
One challenge is how we train a multi-task model from diverse datasets, where each may only contain partial ground-truths for one task.
Here, the problem can be treated within the semi-supervised learning (SSL) paradigm where the labeled data in one dataset is fully supervised and the unlabeled data in other datasets is used in a semi-supervised manner.
Specifically, we use the adversarial learning \cite{goodfellow2014gan} in the structured prediction space for semi-supervised training and iterate a similar scheme on all datasets and tasks.
As such, all tasks are accomplished in the MTL setting.
The second challenge arises from the domain discrepancy across diverse datasets as the data distributions often vary significantly.
Therefore, it is important to align features across domains to learn a model that generalizes to different datasets.
However, unlike the common setting using adversarial alignment \cite{chen2019adversarial,thai2018adaptseg} that considers two domains to distinguish the real-fake distributions, we have multiple prediction distributions from diverse datasets.
Therefore, we propose a domain-aware discriminator structure and analyze various learning modes to mitigate the domain discrepancy. The strategy differentiates the prediction distributions from multiple domains, which in turn enforces the generator to produce more plausible results to confuse the discriminator.

In practice, we focus on two fundamental yet challenging pixel-wise prediction tasks: semantic segmentation \cite{long2015fcn,chen2018deeplab} and depth/height estimation \cite{eigen2014depth,godard2017unsupervised}.
These two tasks learn the semantic and geometric properties for scene understanding, where their correlation has been explored by joint training \cite{zhang2018joint,atapourabarghouei2019veritatem}.
We validate the effectiveness of our SemiMTL method in the challenging autonomous driving and remote sensing scenes, under various settings including the cross-city, cross-dataset, and cross-domain scenarios.

The contributions of this work are summarized as follows: 1) We propose a multi-task learning setting that leverages supervisory signals from diverse datasets in a semi-supervised paradigm; 2) We introduce a domain-aware adversarial learning approach to address the domain discrepancy problem during training across different datasets; 3) We demonstrate the effectiveness of our proposed method for semi-supervised multi-tasking across datasets in challenging street view and remote sensing scenarios.
	
\section{Related Work}
\label{sec:ralated_work}
\noindent\textbf{Multi-task Learning.} MTL has been widely used in vision tasks, such as instance segmentation \cite{goel2021quadronet}, semantic segmentation \cite{liu2018ern, zhen2020joint}, and face analysis \cite{huang2021age}. 
As discussed in \cite{sebastian2017overview}, MTL is typically conducted with either hard or soft parameter sharing of hidden layers in the context of deep learning \cite{misra2016crossstitch,kokkinos2017ubernet,xu2018PADNet,maninis2019attentive,zhang2019pattern, zhou2020pattern, vandenhende2020mti}.
On the other hand, several approaches explore to adaptively calibrate the relative losses of different tasks instead of a naive weighted summation \cite{chen2018gradnorm,sener2018multi,kendall2018multi, yu2020gradient, borse2021inverseform, vasu2021instance}. 
MTL can also be integrated with other learning paradigms, including unsupervised \cite{zou2018df}, self-supervised \cite{ren2018crossdomain}, and transfer learning \cite{zamir2018taskonomy,ramirez2019adat, chavhan2021adaatdt}, to either improve the performance of supervised MTL via additional information or use MTL to facilitate other paradigms \cite{zhang2017survey}. 
A more comprehensive discussion of deep MTL methods can be found in \cite{vandenhende2020mtlsurvey}. 
Note that this study aims to solve a new learning paradigm for MTL rather than designing specific MTL architectures, as our method is compatible with other general MTL networks.

A few semi-supervised MTL methods have been developed \cite{lu2014semimtl,cheng2016semimultimodal}, but do not address the challenging pixel-level tasks 
to train models across different datasets with the absence of ground-truths. 
While some efforts have been made \cite{li2018lwf,kokkinos2017ubernet,pentina2017multi,ramirez2019adat, chavhan2021adaatdt}, these approaches do not conduct synchronous MTL on different datasets.
To optimize jointly with the new labeled task, \cite{li2018lwf} preserves the models trained on old tasks to provide pseudo-ground-truth for these unlabeled tasks, where the joint training strategy can be seen as an upper-bound of their performance.
The UberNet \cite{kokkinos2017ubernet} model is proposed to update network parameters after observing sufficient samples to simulate asynchronous joint training.
However, it does not account for domain gaps across diverse datasets. 
On the other hand, several recent approaches \cite{pentina2017multi,ramirez2019adat, chavhan2021adaatdt} mainly tackle knowledge transfer across tasks rather than across datasets with partial annotations. 

\noindent\textbf{Semi-supervised Learning.} SSL methods leverage the vast amount of unlabeled data for classification and regression problems. 
Perturbation-based methods \cite{oliver2018realistic,luo2018smooth} aim to utilize a teacher model to teach a student module whose predictions should be consistent. 
Similarly, several approaches exploit the fusion strategy by stochastic feature selection \cite{lee2019ficklenet} or learning from multiple regressors \cite{li2017learning}. 
Another line of research in SSL encourages the model to generate confident predictions on unlabeled data, $\emph{e.g.}$, entropy minimization \cite{yves2005entropymin} and pseudo-labeling \cite{hung2018semiseg}.
Auxiliary tasks can also be applied for SSL to integrate supervised and un-/weakly-supervised learning \cite{rasmus2015sslladder,hong2015decoupledseg}. 
More recently, several models in the adversarial setting \cite{goodfellow2014gan} have been developed to either generate realistic samples for better discrimination \cite{souly2017semissgan} or distinguish directly the prediction for better generation \cite{hung2018semiseg,lai2017semiof}. 
Nevertheless, the methods based on adversarial learning are not designed within the MTL framework. 

    \begin{figure*} [!tbp]
    \vspace{-4mm}
    	\begin{center}
    		\includegraphics[width=0.8\textwidth]{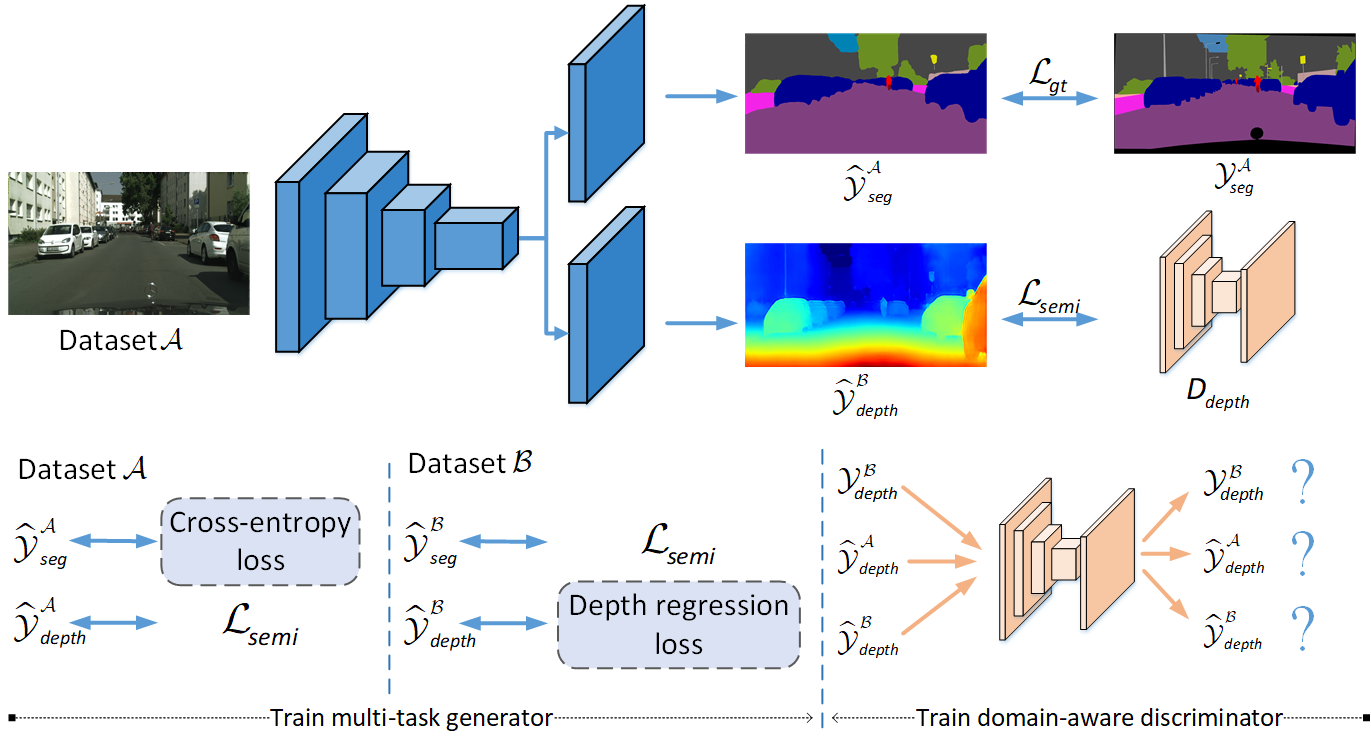}
    	\end{center}
    	\vspace{-4mm}
    	\caption{\textbf{Overview of the proposed SemiMTL method.}  We design the SemiMTL algorithm based on an adversarial learning framework. 
    	During the training process within each dataset, we leverage the ground-truth to supervise the labeled tasks and construct one task-specific discriminator for each unlabeled task to provide semi-supervisory signals. 
    	We optimize the MTL network simultaneously for all task branches over all datasets.
    	The discriminators can be updated after observing the ground-truth and predictions from different datasets to learn the domain knowledge.}
    	\label{fig:framework}
    	\vspace{-4mm}    	
    \end{figure*}

\noindent\textbf{Dense Prediction for Scene Understanding.} Scene understanding involves a group of regression and classification tasks, and we discuss the most related work for semantic segmentation and depth estimation. 
Semantic segmentation can be treated as a pixel-wise classification problem tackled via deep models, such as the FCN \cite{long2015fcn} and Deeplab \cite{chen2018deeplab} networks. 
The recent methods mainly emphasize on learning and assembling features from multiple scales \cite{zhao2017pspnet,he2019dynamic} or multiple layers \cite{bilinski2018dense, takahashi2021densely}, or leveraging global context information \cite{hung2017gce,zhang2018context}.
Similarly, deep models have been successfully applied to depth estimation  \cite{eigen2014depth,godard2017unsupervised}, and numerous algorithms have been developed through
supervised \cite{xu2018structured},
semi-/self-supervised \cite{ramirez2018geometry,godard2019digging},
unsupervised \cite{godard2017unsupervised,nath2018adadepth} ,
and multi-tasking \cite{lai19bridging,chi2021feature} settings.
Since these two tasks are closely related to learning the semantics and geometry of scenes, it is of great interest to accomplish them in a unified framework \cite{zhang2018joint,atapourabarghouei2019veritatem}.
	
\section{SemiMTL}
\label{sec:semimtl}
In this work, we treat the labeled and unlabeled tasks as supervised and semi-supervised problems respectively during the training process within one dataset, which allows us to leverage unlabeled data to further train the task branches without annotations.
We propose the SemiMTL method within the adversarial learning framework \cite{hung2018semiseg,thai2018adaptseg}, where the discriminator and adversarial loss play the role of training signals when the annotation is not available for the training samples. 
In addition, we present a domain-aware structure for the discriminator and analyze different alignment patterns to address the domain gap in multi-tasking across datasets. 
This alleviates the domain discrepancy issue while leveraging the supervision signals from diverse datasets.
In the remainder of this section, we formulate the proposed SemiMTL framework for dense prediction tasks in scene understanding.

\subsection{Approach Overview}
\label{sec:semimtl_approach}
\noindent\textbf{Problem Definition.} We start with considering a typical MTL problem over an input space $\mathcal{X}$ and a collection of task output spaces $\left\{\mathcal{Y}_t\right\}_{t\in\mathcal{T}}$, where $\mathcal{T}$ is the total task set. Given such a dataset, we wish to learn the prediction model per task as $G_t(x;\theta^{sh},\theta^{t}): \mathcal{X}\rightarrow\mathcal{Y}_t$, where $\theta^{sh}$ are the shared parameters among tasks and $\theta^{t}$ are the task-specific parameters. 
In this work, we address a different MTL setting for training across datasets, where each dataset only contains annotations for partial tasks.
Therefore, we extend the setting in \cite{kokkinos2017ubernet,zamir2018taskonomy} and denote the input spaces $\left\{\mathcal{X}^k\right\}_{k\in\mathcal{S}}$, where $\mathcal{S}$ is the set of all datasets.
Here we assume to have two datasets ($\mathcal{A}$ and $\mathcal{B}$) and two tasks ($\mathcal{T}_{seg}$ and $\mathcal{T}_{depth}$) as semantic segmentation and depth estimation. However, each dataset only has some supervision, $\emph{i.e.}$, $\mathcal{Y}^\mathcal{A}_{seg}$ for task $\mathcal{T}_{seg}$ in dataset $\mathcal{A}$ and $\mathcal{Y}^\mathcal{B}_{depth}$ for task $\mathcal{T}_{depth}$ in dataset $\mathcal{B}$, as shown in Figure \ref{fig:framework}.

\vspace{1mm}
\noindent\textbf{Baseline: Joint Training.} We first apply a joint training baseline method \cite{kokkinos2017ubernet} that iteratively explores each dataset and updates the model ($\emph{i.e.}, \theta^{sh}$ and $\theta^{t}$) after observing sufficient annotated samples for each task. 
However, leveraging only partial annotations from each dataset may produce bias in the shared encoder as it does not observe gradients from unlabeled data.
As a result, the model may perform well on the labeled tasks while generalizing poorly on the unlabeled tasks in each dataset.

\noindent\textbf{Proposed Method.} To address the above-mentioned issue, we propose to train the model on one domain $\mathcal{X}^k$ by minimizing the supervised loss for labeled task $\mathcal{T}_t$ with annotated samples $(x^k,y^k_t)$ in $(\mathcal{X}^k,\mathcal{Y}^k_t)$, as well as the semi-supervised loss for unlabeled tasks $\mathcal{T}\setminus\mathcal{T}_t$ with identical samples $x^k$ in $\mathcal{X}^k$ that do not have corresponding annotations.
To consider all the input domains $\mathcal{X}$, we iteratively apply the above training scheme over each dataset $\mathcal{X}^k$ to fully leverage the supervisory signals for each task.

With this formulation, our model is able to optimize all the task-specific decoders $\left\{\mathit{F}_t; \theta^t \right\}_{t\in\mathcal{T}}$ simultaneously with supervisions either from the supervised loss or the semi-supervised one on unlabeled data. 
Therefore, the shared encoder $\left\{\mathit{E}; \theta^{sh} \right\}$ can also update with gradients accumulated from the supervision of all tasks on both labeled and unlabeled data, which avoids the bias only on labeled data.
We will describe the details about our proposed framework and semi-supervised loss in the following sections.

\subsection{Objective Function}
\label{sec:semimtl_objective}
We formulate our SemiMTL problem as an adversarial learning framework, which consists of two modules: the generator $\mathit{G}$ and the discriminators $\left\{\mathit{D}_t\right\}_{t\in\mathcal{T}}$. The generator $\mathit{G}$ is a multi-task network that contains a shared encoder $\mathit{E}$ parameterized by $\theta^{sh}$ and task-specific decoders $\left\{\mathit{F}_t\right\}_{t\in\mathcal{T}}$ parameterized by $\theta^{t}$. 

\noindent\textbf{Typical MTL Loss.}  Given an input image $x\in\mathbb{R}^{H\times W \times 3}$, the typical MTL loss function is defined by the task-specific supervised loss $\mathcal{L}_{gt}^t$ for task $t$, with a weight $w_t$ to balance the loss functions among tasks:

    \begin{equation}
	\label{eq:mtl}
		\begin{aligned}
			\mathcal{L}_{mtl} = \sum_{t=1}^{|\mathcal{T}|} w_t\mathcal{L}_{gt}^t(F_t(E(x)), y_t),
		\end{aligned}
	\end{equation}
where $y_t$ is the ground-truth for task $t$.

\noindent\textbf{Discriminator Module.} In our setting, each dataset may not contain the label for all the tasks, indicating that the task-specific branches cannot be supervised by unlabeled data via Eq. \eqref{eq:mtl}.
Here, our goal is to accomplish all tasks to simultaneously update both the shared encoder and task-specific decoders using both labeled and unlabeled data. 
To this end, we utilize adversarial learning to construct a semi-supervised objective for the data without ground-truths. 
%
Our approach is motivated by the observation that the output space is structured in dense prediction tasks such as semantic segmentation and depth estimation \cite{hung2018semiseg,thai2018adaptseg}. 
For example, the street-view images might have significant differences in appearance, but their outputs share many similarities such as spatial layout and local context.

In \cite{thai2018adaptseg}, they introduce a discriminator to distinguish whether the distribution is from the ground-truth or the prediction of unlabeled data.
Differently, our method deals with the SemiMTL setting that contains labeled data from one domain and unlabeled data from other domains.
Thus, we introduce a discriminator module that can tell which domain that the prediction comes from, $\emph{i.e.}$, either the ground-truth or the prediction of the domain $\mathcal{X}^k$.
Specifically, we first forward the input image $x$ to the generator network $G = \{E; F_t\}$ and produce the task-specific prediction $\hat{y}_t = F_t(E(x))$ for task $t$, which is then taken as the input to the discriminator. 
We minimize the cross-entropy loss $\mathcal{L}_{D}^t$ for the task-specific discriminator $\mathit{D_t}$:
    \begin{equation}
	\label{eq:disc}
		\begin{aligned}
			\mathcal{L}_{D}^t &= \mathcal{L}_{ce}(z_t, c)
			   = - \sum_{h,w} c \log(D_t(z_t)^{(h,w,c)}),
		\end{aligned}
	\end{equation}
where $c$ is the one-hot domain label and $z_t$ denotes the input to the discriminator, which could be the ground-truth $y_t$ ($c=0$) or the prediction $\hat{y}_t$ from the $c$-th dataset (domain).
%
In this paper, $c$ is a 3-dimensional one-hot vector, in which a three-way classifier is utilized in the discriminator to tell whether the input is from the ground-truth or the prediction from dataset $\mathcal{A}$ or $\mathcal{B}$.
%
    \begin{figure} [!tb]
        \begin{center}
            \includegraphics[width=0.45\textwidth]{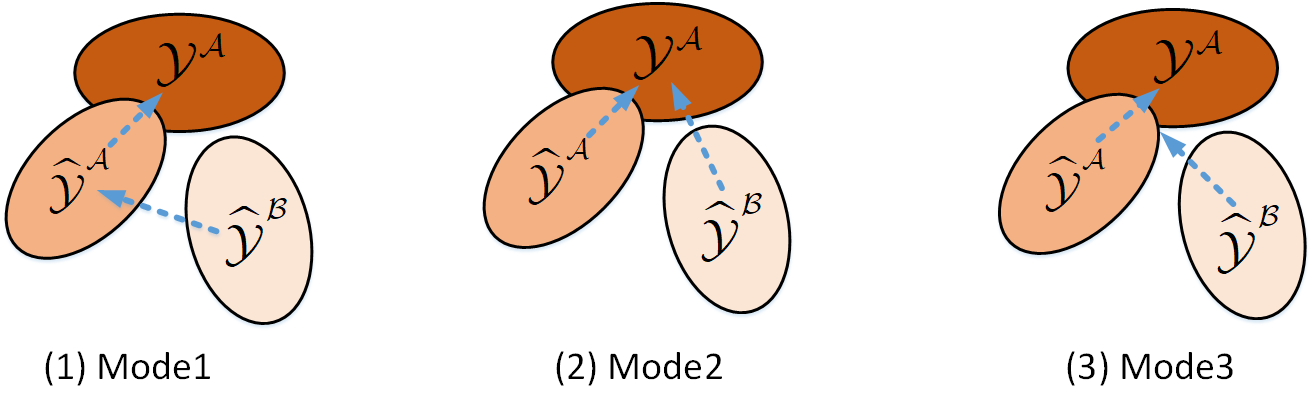}
    	\end{center}
    	\vspace{-3mm}
    	\caption{\textbf{Illustration of alignment strategies.} In our setting, assume only dataset $\mathcal{A}$ contains the annotations $\mathcal{Y}^{\mathcal{A}}$ for task $\mathcal{T}_{seg}$. The predictions $\hat{\mathcal{Y}}^{\mathcal{A}}$ and $\hat{\mathcal{Y}}^{\mathcal{B}}$ are obtained from the labeled dataset $\mathcal{A}$ and unlabeled dataset $\mathcal{B}$ respectively. For the alignment of this task, we directly enforce $\hat{\mathcal{Y}}^{\mathcal{A}}$ as close to ground-truth distribution as possible, while align $\hat{\mathcal{Y}}^{\mathcal{B}}$ with different strategies, that is, to the distributions of (1) prediction $\hat{\mathcal{Y}}^{\mathcal{A}}$, (2) ground-truth $\mathcal{Y}^{\mathcal{A}}$, or (3) their intersection $\hat{\mathcal{Y}}^{\mathcal{A}} \cup \mathcal{Y}^{\mathcal{A}}$.}
    	\label{fig:adversarial_modes}
    	\vspace{-3mm}    	
    \end{figure}

\noindent\textbf{Adversarial Loss.}
Based on the discriminator, our goal is to provide training signals for unlabeled data and enforce the prediction to be close to the ground-truth distribution.
%
However, it is not trivial to directly apply conventional adversarial alignment \cite{hung2018semiseg,thai2018adaptseg} as our setting involves predictions from multiple datasets. 
Here, we investigate several alignment strategies that can achieve the desired goal, as shown in Figure \ref{fig:adversarial_modes}. 
Suppose the dataset $\mathcal{A}$ is labeled for task $t$ but not for the dataset $\mathcal{B}$,
we note that there are three types of distributions for each task: the ground-truth $y_t^\mathcal{A}$, the prediction from the labeled dataset $\hat{y}_t^{\mathcal{A}}$, and the prediction from the unlabeled dataset $\hat{y}_t^{\mathcal{B}}$. 
We can directly align the distribution of $\hat{y}_t^{\mathcal{A}}$ to $y_t^{\mathcal{A}}$ as they are from the same domain $\mathcal{A}$ (intra-domain loss):
    \begin{equation}
	\label{eq:adv_intra}
		\mathcal{L}_{intra}^t = - \sum_{h,w} \log(D_{t}(\hat{y}_t^{\mathcal{A}})^{(h,w,0)}),
    \end{equation}
where label 0 indicates the ground-truth.
For the inter-domain loss, we exploit different training modes and construct the corresponding loss function.
Here, we can choose to align the distribution of prediction $\hat{y}_t^{\mathcal{B}}$ from the unlabeled data to the one from the labeled domain, $\emph{i.e.}$,  $\hat{y}_t^{\mathcal{A}}$ (labeled as 1) as Mode 1:
    \begin{equation}
	\label{eq:adv_inter1}
		\mathcal{L}_{inter}^t = - \sum_{h,w} \log(D_{t}(\hat{y}_t^{\mathcal{B}})^{(h,w,1)}),
    \end{equation}
or the ground-truth $y_t^\mathcal{A}$ as Mode 2:
    \begin{equation}
	\label{eq:adv_inter2}
		\mathcal{L}_{inter}^t = - \sum_{h,w} \log(D_{t}(\hat{y}_t^{\mathcal{B}})^{(h,w,0)}),
	\end{equation}
or the joint distribution of $y_t^\mathcal{A}$ and $\hat{y}_t^{\mathcal{A}}$ as Mode 3:
    \begin{equation}
	\label{eq:adv_inter3}
		\mathcal{L}_{inter}^t = - \sum_{h,w} \log(1-D_{t}(\hat{y}_t^{\mathcal{B}})^{(h,w,2)}),
	\end{equation}
where label 2 denotes the prediction of unlabeled data $\hat{y}_t^{\mathcal{B}}$.
%
Hence we define the semi-supervised loss for task $t$ as
    \begin{equation}
	\label{eq:semi}
		\mathcal{L}_{semi}^t = \lambda_{intra}\mathcal{L}_{intra}^t + \lambda_{inter}\mathcal{L}_{inter}^t,
	\end{equation}
where $\lambda_{intra}$ and $\lambda_{inter}$ indicate the weight for the intra- and inter-domain adversarial losses, respectively.

We denote the SemiMTL framework applying these alignment modes as SemiMTL (M1), SemiMTL (M2), and SemiMTL (M3), respectively. We utilize the SemiMTL (M2) mode as our default implementation. Their effect on the performance of tasks is illustrated in Section \ref{sec:exp_ablation}.
Note that we utilize the same adversarial training scheme to the task $\mathcal{T}_{seg}$ and $\mathcal{T}_{depth}$, which is also applicable to other similar tasks.
    
    \begin{algorithm}[!tbp]
    \scriptsize
    \caption{Training procedure of the SemiMTL method.}
    \label{alg:semimtl_procedure}
        \begin{algorithmic}
        \For{iteration $i=1$ to $N$}
            \For{dataset $k \in \left\{1, ..., K\right\}$}
                \State \{Construct mini-batch\}
                \For{task $t = 1, ..., m$ with ground-truth}
                    \State \{calculate gradients for $\mathit{E}$ and $\mathit{F}_t$\}
                    \State $\textbf{L}^t_G \gets w_t \mathcal{L}^t_{gt} + \lambda_{intra} \mathcal{L}_{intra}^t $
                    \State \{calculate gradients for $\mathit{D}_t$\}
                    \State $\textbf{L}^{t,k}_D \gets \mathcal{L}^t_D(y_t,0) + \mathcal{L}^t_D(\hat{y}_t,k)$
                \EndFor
                \For{task $t = m+1, ..., |\mathcal{T}|$ without ground-truth}
                    \State \{calculate gradients for $\mathit{E}$ and $\mathit{F}_t$\}
                    \State $\textbf{L}^t_G \gets \lambda_{inter} \mathcal{L}^t_{inter}$
                    \State \{calculate gradients for $\mathit{D}_t$\}
                    \State $\textbf{L}^{t,k}_D \gets \mathcal{L}^t_D(\hat{y}_t,k)$
                \EndFor
                \State \{Freeze $\left\{\mathit{D}_t\right\}$, and update $\mathit{G}$ with\}
                \State $\textbf{L}_G = \sum^{|\mathcal{T}|}_{t=1} \textbf{L}^t_G$
            \EndFor
            \State \{Freeze $\mathit{G}$, and update $\left\{\mathit{D}_t\right\}$ with\}
            \State $\textbf{L}^t_D = \sum^K_{k=1} \textbf{L}^{t,k}_D$
        \EndFor
        \end{algorithmic}
    \end{algorithm}        

\subsection{Optimization}
\label{sec:semimtl_optimization}
We apply the synchronous SGD for joint training \cite{kokkinos2017ubernet} as the baseline training method, where we extract mini-batches from every dataset iteratively and optimize all parameters synchronously using \eqref{eq:mtl} after observing labeled samples for each task. 
With the baseline training approach, we construct our SemiMTL model, where the unlabeled tasks guided by the discriminators can be optimized simultaneously with the labeled ones. 
The main steps of this process are summarized in Algorithm \ref{alg:semimtl_procedure}.
Within each training iteration, we minimize the overall objective functions for the generator:
    \begin{equation}
	\label{eq:overall_gen}
		\textbf{L}_G = \sum^{|\mathcal{T}|}_{t=1} w_t\mathcal{L}^t_{gt} +  \mathcal{L}_{semi}^t,
	\end{equation}
and for each discriminator $\mathit{D}_t$:
    \begin{equation}
	\label{eq:overall_disc}
		\textbf{L}^t_D = \mathcal{L}^t_D(y_t,0) + \sum^K_{k=1}\mathcal{L}^t_D(\hat{y}_t,k),
	\end{equation}
where $K$ is the number of datasets/domains.
The SemiMTL model is iteratively trained in a way similar to the GAN \cite{goodfellow2014gan} method: the discriminator $\mathit{D}_t$ aims to classify the ground-truth/predictions from different domain distributions, while the generator $\mathit{G}$ attempts to fool $\mathit{D}_t$ by producing predictions that are as indistinguishable to the ground-truth as possible.

\subsection{Network Architecture}
\label{sec:semimtl_network}
\noindent\textbf{Encoder and Decoder Networks.} The proposed SemiMTL framework can integrate any type of deep MTL architectures. Here we adopt the commonly utilized encoder-decoder MTL model that consists of a shared encoder coupled with two task-specific decoders to estimate segmentation and depth tasks. 
We leverage the ResNet101 \cite{he2016resnet} backbone for the shared encoder to obtain deep feature representations, which are passed to two parallel branches for independent task decoding. 
The segmentation decoder is built upon the PSP module \cite{zhao2017pspnet} to increase contextual information for semantics, followed by a softmax layer to predict semantic classes. 
The depth decoder is constructed with several convolutional layers and up-sampling operations to produce detailed depth features and a regression layer to estimate depth. 
Finally, we apply an up-sampling layer to the output maps for both tasks to match the input image size. 
To optimize the network, we adopt the cross-entropy loss for semantic segmentation and the BerHu loss \cite{laina2016deeper} for depth estimation.

\noindent\textbf{Discriminator Networks.} The structure of the discriminator network is similar to that in \cite{radford2015unsupervised}.
It consists of 5 convolution layers with $4\times4$ kernel and $\{64, 128, 256, 512, K\}$ channels in the stride of 2. 
The first four convolution layers are all followed by a spectral normalization layer \cite{miyato2018spectral} to stabilize the training process and a leaky ReLU \cite{maas2013relu} unit parameterized by 0.2. 
We implement an up-sampling layer to transform the output to the input size. 
The discriminators for both tasks share the same architecture except for the input layer which takes the segment and depth maps respectively.

\section{Experimental Results and Analysis}
\label{sec:experiments}
To demonstrate the effectiveness of the SemiMTL method, we carry out experiments on several publicly available datasets for scene understanding, including the ISPRS Potsdam and Vaihingen \cite{ISPRS_Datasets} remote sensing datasets, and the real-world Cityscapes \cite{cordts2016cityscapes} and synthetic Synscapes \cite{wrenninge2018synscapes} street-view datasets. 
%
%
In particular, we evaluate the algorithms in three scenarios, including the cross-city, cross-dataset, and cross-domain settings with various datasets.
In the following, we will describe the experimental details.

\subsection{Experimental Setup}
\label{sec:exp_setup}

\noindent\textbf{Datasets.} The Cityscapes dataset contains high-resolution outdoor images for urban scene understanding, which is collected from 50 diverse cities. It is annotated with pixel-wise semantic labels, associated with pre-computed disparity maps that can serve as pseudo depth ground-truth. 
The Cityscapes-depth dataset is an additional train-extra set of Cityscapes with disparity ground-truths, which is collected from different cities with the Cityscapes. 
The Synscapes dataset is generated by photorealistic rendering techniques to parse synthetic street scenes, containing 25K RGB images together with accurate pixel-wise class and depth annotations. Since this dataset does not provide an official split, we take the consequent 20K/1K/4K samples as our training/validation/test sets respectively.
We estimate the inverse depth to represent points at an infinite distance like the sky as zero. As the images of these datasets are of high resolution, we resize the images to 512$\times$1024 for experiments. 

The ISPRS Potsdam and Vaihingen datasets were acquired by flight campaigns over German cities, accompanied by digital images, semantic labels, and digital surface model (DSM) height data.
The digital images were captured by the airborne color-infrared camera in different channels: near-infrared (NIR)/infrared (IR), red (R), and green (G).
The DSM data was acquired by LiDAR and the normalized DSM (nDSM) data was also made available, which is normalized to the range [0,1] in our experiments.
The images in Potsdam and Vaihingen are composed of the IRRG bands at a ground sample distance (GSD) of 5cm and the NIRRG bands at a GSD of 9cm, respectively.
We follow the official split for training and testing set as in \cite{ISPRS_Datasets}.
In our experiments, we resize the image data in Potsdam to the GSD of 9cm to match with the Vaihingen dataset.
We then extract patches of size 512$\times$512 from the raw high-resolution images using a $50\%$ overlapped sliding window along both the row and column.

\noindent\textbf{Evaluation Metrics.} For the evaluation of segmentation, we use the mean pixel accuracy (pAcc) and mean Intersection over Union (mIoU) metrics. The pAcc indicates the total accuracy of pixels regardless of classes while the mIoU is computed by averaging the Jaccard scores over all predicted categories. To evaluate the depth task, we adopt several quantitative metrics following \cite{eigen2014depth,godard2017unsupervised}, 
including (a) abs relative error (AbR), (b) root mean squared error (RMSE), and (c) accuracy with thresholds: $\%$ of $\hat{y}_n$ s.t. $\max(\frac{\hat{y}_n}{y_n}, \frac{y_n}{\hat{y}_n})=\delta_i < 1.25^i$ $(i\in[1, 2, 3])$, where $\hat{y}_n$ and $y_n$ denote the prediction and ground-truth of depth at the $n$-th pixel.
We also measure the multi-task performance $\Delta M$ \cite{maninis2019attentive}, i.e. the average per-task performance gain of multi-task model $m$ compared with the single-task baseline $b$ :
    \begin{equation}
	\label{eq:mtl_metric}
		\Delta M = \frac{1}{|\mathcal{T}|} \sum_{t=1}^{|\mathcal{T}|} (-1)^{l_t} (M_{m,t}-M_{b,t})/M_{b,t},
	\end{equation}
where $M$ indicates the representative measure for each task and we adopt the mIoU and RMSE metric as in \cite{maninis2019attentive}.
$l_t = 1$ if a lower $M$ means a better performance, and 0 otherwise. 

\noindent\textbf{Implementation Details.} The SemiMTL method is implemented with PyTorch using Nvidia Titan RTX GPUs. We initiate the encoder backbone parameters with the ResNet101 \cite{he2016resnet} model pre-trained on ImageNet, and the decoders and discriminators are randomly initialized. We perform the data augmentation on the fly following \cite{zhang2018context} and fix the crop size during the training process. The MTL network is trained by the standard SGD optimizer \cite{bottou2010sgd} with momentum 0.9 and weight decay $10^{-4}$. The learning rate is initialized by 0.01 and decreased using the polynomial decay with power 0.9. We adopt the Adam optimizer \cite{kingma2014adam} for training the discriminators with learning rate as $10^{-4}$ and momentum as (0.9, 0.99). In all experiments, we fix the task weights as $w_{seg}=1.0$ and $w_{depth}=0.01$ in the MTL loss and set $\lambda_{intra}=0.001$ and $\lambda_{inter}=0.0001$ to balance the semi-supervised adversarial losses. We use the same hyper-parameters among all methods for fair comparisons.

\subsection{Evaluation of SemiMTL Framework}
\label{sec:exp_overall}

We compare the experimental results of the SemiMTL approach with different baselines. 
We first build the models for each task with identical encoder structure and task-specific decoder head, termed as single task learning (STL). Then we utilize the joint training algorithm \cite{kokkinos2017ubernet}, named joint task learning (JTL), to train the MTL model across datasets. 
We also apply a domain adaptation algorithm \cite{thai2018adaptseg} to both STL and JTL schemes, which does not consider the prediction distributions from different domains.
Extensive experiments demonstrate the effectiveness of our method to leverage semi-supervised information during multi-tasking across datasets.

\begin{table}[!t]
\newcommand{\tabincell}[2]{\begin{tabular}{@{}#1@{}}#2\end{tabular}}  
\centering
\caption{\textbf{Quantitative results on the Cityscapes dataset.} We train all methods with the training and train-extra sets, and evaluate them on the validation set for both tasks. 
The cyan metrics indicate lower is better while pink ones mean higher is better.}
\label{tab:across_cities}
\renewcommand\arraystretch{1.0}	
\setlength{\tabcolsep}{1pt}
\scriptsize	
\begin{tabular}{ c p{0.8cm}<{\centering} p{0.8cm}<{\centering} p{0.8cm}<{\centering} p{0.8cm}<{\centering} p{0.7cm}<{\centering} p{0.7cm}<{\centering} p{0.7cm}<{\centering} p{1.0cm}<{\centering}}
\toprule
\multirow{2}{*}{Method}  &\multicolumn{2}{c}{Segmentation}                       &\multicolumn{5}{c}{Depth}  &\multicolumn{1}{c}{MTL}\\ 
\cmidrule(lr){2-3} \cmidrule{4-8} \cmidrule(lr){9-9}
    &pAcc &mIoU   &\cellcolor{mycyan}AbR &\cellcolor{mycyan}RMSE &\cellcolor{mypink}{$\delta_1$}    &\cellcolor{mypink}$\delta_2$  &\cellcolor{mypink}$\delta_3$
    &$\Delta M(\%)$\\
\midrule
STL\_Seg                       &94.8   &71.4   &-      &-     &-     &-     &-      &0.0 \\ 
STL\_Depth                     &-      &-      &0.414  &6.744 &67.6  &84.5  &92.0   &0.0 \\
JTL \cite{kokkinos2017ubernet} &94.8   &71.4   &0.329  &5.469 &76.6  &91.2  &95.7   &+9.4 \\
SemiMTL            &\textbf{94.9}   &\textbf{71.9}   &\textbf{0.287} &\textbf{5.234}  &\textbf{79.3}  &\textbf{92.6}  &\textbf{96.3}  &\textbf{+11.5}\\
\bottomrule
\end{tabular}
\vspace{-4mm}
\end{table}

\noindent\textbf{Across Cities.} 
In this setting, we conduct experiments on the Cityscapes and Cityscapes-depth datasets where the former and latter only provide segmentation and depth ground-truths, respectively. 
They are captured from different European cities at different seasons, which can verify our method in a small domain gap scenario. 
We train the methods on the training sets of two datasets while evaluating them on the Cityscapes validation set containing the labels for both tasks. We fix the crop size as 256$\times$256 during the training step.

Table \ref{tab:across_cities} shows the evaluation results of our proposed algorithm against baseline methods, where our method achieves the best performance on both tasks. 
It is worth noting that compared with the separate training of each task, the joint training of both tasks achieves identical results for the segmentation task but obtains significant improvements on the depth task. 
The results indicate that the high-level segmentation task facilitates more on the low-level depth task. 
The proposed SemiMTL method improves further on all tasks compared with the JTL scheme. 
%

\begin{table}[!t]
\newcommand{\tabincell}[2]{\begin{tabular}{@{}#1@{}}#2\end{tabular}} 
\centering
\caption{\textbf{Quantitative results on the Potsdam and Vaihingen datasets.} We train all the methods with the segmentation ground-truth from Potsdam and depth ground-truth from Vaihingen, while evaluating the performance of each task on both datasets to validate whether the above trained models generalize well across datasets.}
\label{tab:across_datasets}
\renewcommand\arraystretch{1.0}	
\setlength{\tabcolsep}{1pt}
\scriptsize	
\begin{tabular}{ c | c  p{0.7cm}<{\centering} p{0.7cm}<{\centering} p{0.7cm}<{\centering} p{0.7cm}<{\centering} p{0.6cm}<{\centering} p{0.6cm}<{\centering} p{0.6cm}<{\centering} p{1.0cm}<{\centering}}
\toprule
\multicolumn{2}{c}{\multirow{2}{*}{Method}}    &\multicolumn{2}{c}{Segmentation}     &\multicolumn{5}{c}{Depth} &\multicolumn{1}{c}{MTL}\\
\cmidrule(lr){3-4} \cmidrule{5-9} \cmidrule(lr){10-10}
\multicolumn{2}{c}{}  &pAcc &mIoU   &\cellcolor{mycyan}AbR &\cellcolor{mycyan}RMSE &\cellcolor{mypink}{$\delta_1$}    &\cellcolor{mypink}$\delta_2$  &\cellcolor{mypink}$\delta_3$ 
&$\Delta M(\%)$\\
\midrule
\multirow{6}{*}{\rotatebox{90}{Potsdam}}   
    &STL\_Seg       &89.5   &79.7   &-  &-  &-  &-  &- &0.0\\ 
    &STL\_Depth     &-  &-      &6.926  &4.686  &16.1   &24.8   &33.7 &0.0\\
    &DA\_Depth \cite{thai2018adaptseg} &-  &-  &4.985  &4.677  &28.5  &36.3  &46.4 &-\\
    &JTL \cite{kokkinos2017ubernet}  &90.0   &80.7   &2.517 &4.430  &31.2  &41.7  &52.3 &+3.4\\
    &SemiSD \cite{thai2018adaptseg}  &90.2   &80.8   &2.695 &4.322  &34.1  &44.7  &55.3 &+4.6\\
    &SemiMTL            &\textbf{90.5}   &\textbf{81.4}   &\textbf{2.420}  &\textbf{4.217} &\textbf{38.4}  &\textbf{47.0}  &\textbf{57.3}  &\textbf{+6.1}\\
\midrule
\multirow{6}{*}{\rotatebox{90}{Vaihingen}}   
    &STL\_Seg   &64.3   &42.4   &-      &-      &-      &-      &-  &0.0\\
    &DA\_Seg \cite{thai2018adaptseg}   &68.8   &47.5   &-  &-  &-  &-  &-   &-\\
    &STL\_Depth &-      &-      &1.324  &1.899  &48.7  &68.9  &\textbf{78.5}  &0.0\\ 
    &JTL \cite{kokkinos2017ubernet}  &79.3  &62.2   &1.338  &1.949  &48.8  &68.1  &78.1 &+22.1\\
    &SemiSD \cite{thai2018adaptseg}  &81.4  &63.8   &1.432  &2.088  &49.6  &67.5  &78.1 &+20.4\\
    &SemiMTL            &\textbf{81.6}  &\textbf{64.9}   &\textbf{1.316}     &\textbf{1.802} &\textbf{50.9}  &\textbf{69.4}  &78.4 &\textbf{+29.2}\\
\bottomrule
\end{tabular}
\vspace{-2mm}
\end{table}
    
    \begin{figure} [!tbp]
    \vspace{-2mm}
    	\begin{center}
    		\includegraphics[width=0.48\textwidth]{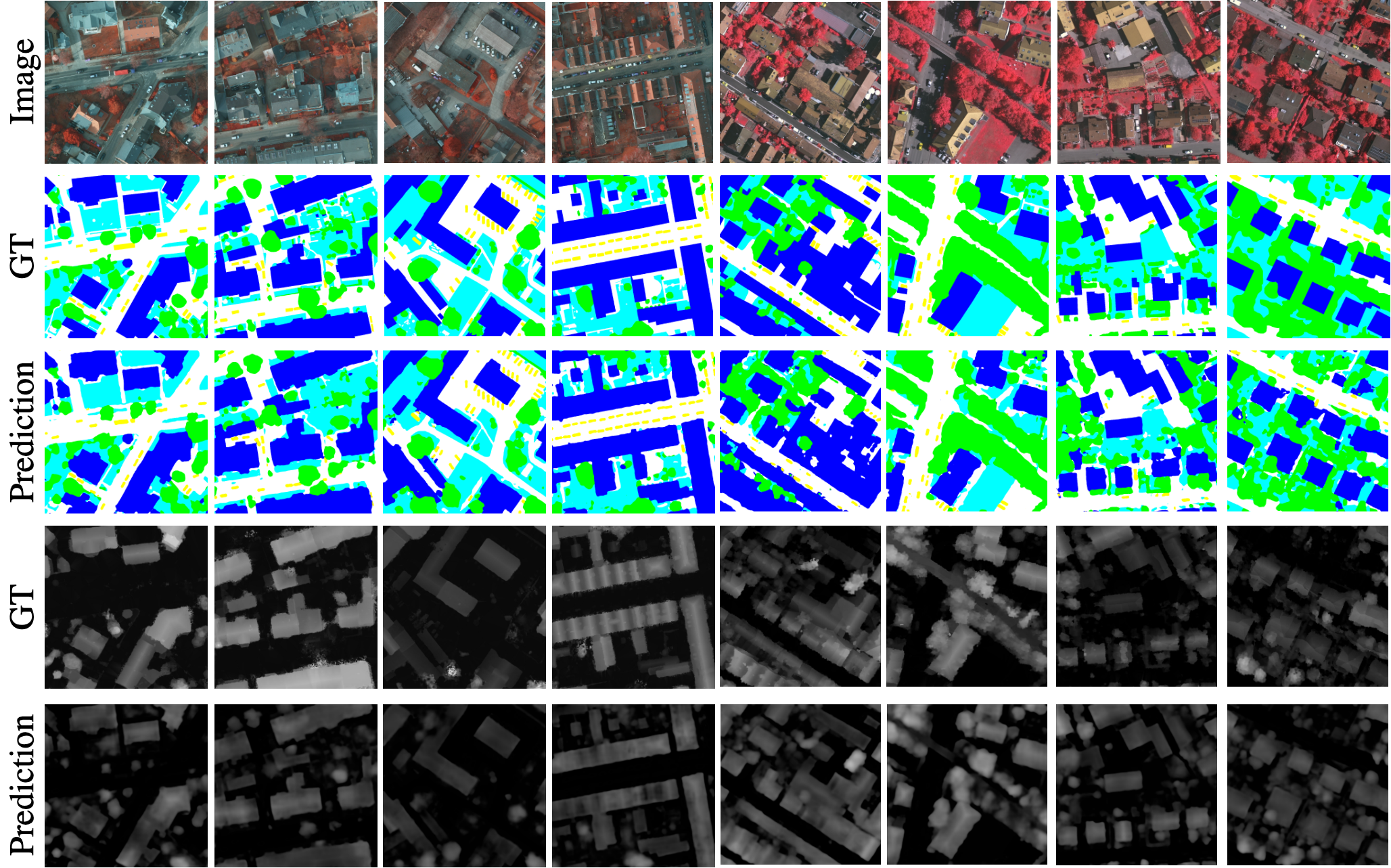}
    	\end{center}
    	\vspace{-3mm}
    	\caption{\textbf{Qualitative examples of the SemiMTL method on the remote sensing scenario.} The first and last four columns are the examples from the Potsdam and Vaihingen, respectively.}
    	\label{fig:pots_vaih}
    	\vspace{-4mm}    	
    \end{figure}

\noindent\textbf{Across Datasets.} We conduct the cross-dataset experiments on the Potsdam and Vaihingen remote sensing datasets, which are collected with significantly different conditions such as imaging sensors, GSD, and even color channel compositions. 
Only the ground-truths of Potsdam segmentation and Vaihingen height are available during the training step. 
We then evaluate each task on the validation set of both datasets to verify the generalization ability, as the performance boost of all tasks on all datasets is preferred rather than improving only the specific tasks with supervision in each dataset. 
We fix the crop size as 384$\times$384 for training.

The quantitative results of each task on both datasets are shown in Table \ref{tab:across_datasets}. 
We adopt the domain adaptation (DA) method \cite{thai2018adaptseg} to the STL of each task, namely, DA$\_$Seg and DA$\_$Depth. This method is originally proposed for semantic segmentation, however, it is also applicable to other dense prediction tasks with structured output space such as depth estimation. 
In the STL experiments, the models perform well on the supervised dataset but obtain poor results on the unsupervised dataset, whereas the DA method is able to improve the performance of every single task. 
The JTL \cite{kokkinos2017ubernet} algorithm leverages more comprehensive supervision from related tasks in each dataset, which can learn useful complementary features in the shared encoder. As such, it achieves significant improvement for the task without supervision in each dataset, $\emph{e.g.}$, the depth task in Potsdam and the segmentation task in Vaihingen.

Based on the above observation, we propose the SemiMTL strategy to further improve the performance of the unsupervised tasks in each dataset.
We again employ the adversarial algorithm \cite{thai2018adaptseg, hung2018semiseg} within our framework, where a task-specific discriminator provides the semi-supervisory signals for each task, termed as SemiSD. With the help of semi-supervision, the unsupervised tasks in each dataset are boosted against the JTL scheme. However, this scheme only transfers the features from the source dataset to target one, namely, aligning the prediction and ground-truth distributions.
As such, we further consider the multi-domain issue and design the domain-aware discriminator to better differentiate the predictions, which in turn forces the generator to produce more realistic results to confuse the discriminator. 
As a result, our SemiMTL method not only improves the semi-supervised tasks with a large margin (4.0$\%$ mIoU gain in Vaihingen and 4.8$\%$ RMSE gain in Potsdam against the JTL baseline) but also achieves performance gains for the fully supervised tasks.
It is worth noting that the performance of unlabeled tasks in each dataset is improved more significantly, which is in line with our motivation to leverage the semi-supervisory signals to improve the unlabeled tasks.

\noindent\textbf{Across Domains.} We further carry out experiments on the real-world Cityscapes and synthetic Synscapes datasets. 
The training on them is much more challenging to suffer from not only larger domain discrepancy across datasets but also differences between real and synthetic scenes. 
We perform the experiments similarly by training with the ground-truths of segmentation in Cityscapes and depth in Synscapes while evaluating each task on the validation set of both datasets.

\begin{figure} [!tbp]
    \vspace{-3mm}
    	\begin{center}
    		\includegraphics[width=0.48\textwidth]{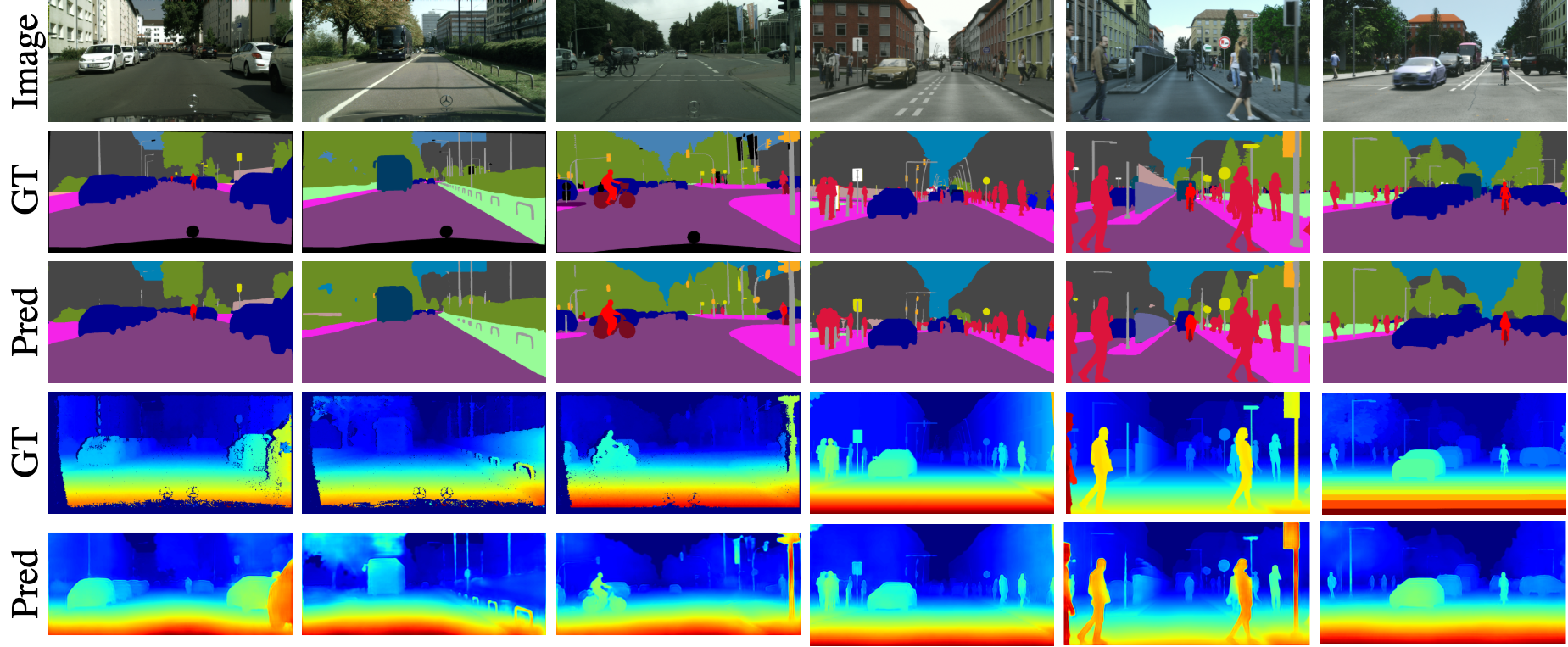}
    	\end{center}
    	\vspace{-3mm}
    	\caption{\textbf{Qualitative examples of the SemiMTL method on the street-view scenario.} The first and last three columns are the examples from the Cityscapes and Synscapes, respectively.}
    	\label{fig:citys_syns}
    	\vspace{-5mm}    	
    \end{figure}

\begin{figure} [!t]
    \vspace{-3mm}
        \begin{center}
            \includegraphics[width=0.48\textwidth]{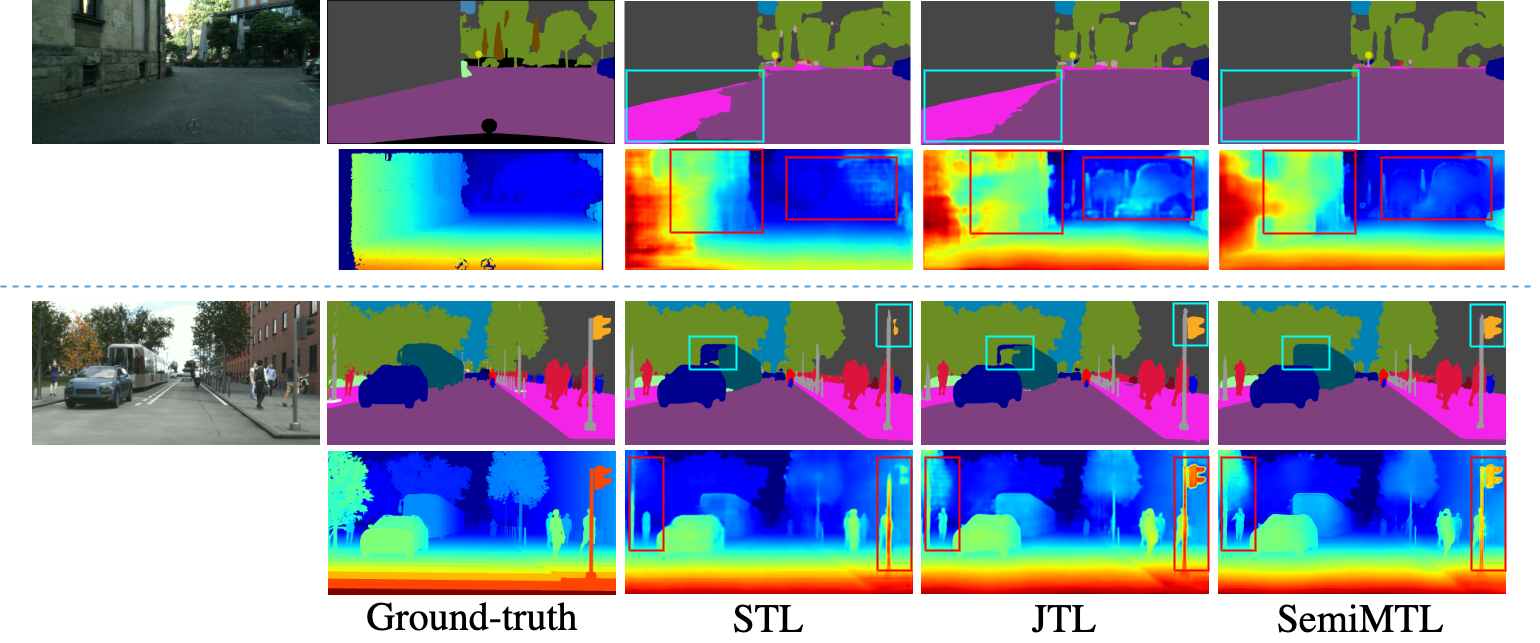}
    	\end{center}
    	\vspace{-3mm}
    	\caption{\textbf{Qualitative comparison for different methods.} The upper and lower examples are from the Cityscapes and Synscapes datasets respectively. 
    	The improvements are highlighted with \textcolor{cyan}{cyan} and \textcolor{red}{red} rectangles for segmentation and depth tasks respectively.}
    	\label{fig:method_comparison}
    	\vspace{-5mm}    	
    \end{figure}

The quantitative results of each task on both datasets are shown in Table \ref{tab:across_domains}.
In the supervised experiments (segmentation of Cityscapes and depth of Synscapes), the JTL method improves the depth task similar to the cross-city setting, while degrading the segmentation mIoU due to the large domain gap among real-synthetic datasets.
In contrast, our SemiMTL framework achieves mIoU performance gain by 1.2$\%$ and 1.8$\%$ against the STL and JTL baselines respectively, and also improves all metrics in the depth task. 
In the semi-supervised experiments (segmentation of Synscapes and depth of Cityscapes), the JTL method improves the performance of both tasks significantly, indicating that the observation of cross-domain samples can help the network to learn more generalized features. 
Compared with the JTL baseline, our SemiMTL framework further facilitates the segmentation task with a performance gain of 3.2$\%$ in mIoU and improves the depth task by 12.6$\%$ in RMSE.
Figure \ref{fig:citys_syns} shows the qualitative results of the proposed SemiMTL algorithm. We also provide the comparisons for different methods in Figure \ref{fig:method_comparison}, which indicates that the proposed method predicts more accurately for segmentation and estimates more sharply along boundaries and smoothly within regions for depth.
%

\begin{table}[!t]
\vspace{-2mm}
\newcommand{\tabincell}[2]{\begin{tabular}{@{}#1@{}}#2\end{tabular}} 
\centering
\caption{\textbf{Quantitative results on the Cityscapes and Synscapes datasets.} We train all methods with the ground-truths of segmentation from Cityscapes and depth from Synscapes, while evaluating each task on both datasets to validate the generalization ability.}
\label{tab:across_domains}
\renewcommand\arraystretch{1.0}	
\setlength{\tabcolsep}{1pt}
\scriptsize	
\begin{tabular}{ c | c  p{0.7cm}<{\centering} p{0.7cm}<{\centering} p{0.7cm}<{\centering} p{0.7cm}<{\centering} p{0.6cm}<{\centering} p{0.6cm}<{\centering} p{0.6cm}<{\centering} p{1.0cm}<{\centering}}
\toprule
\multicolumn{2}{c}{\multirow{2}{*}{Method}}    &\multicolumn{2}{c}{Segmentation}     &\multicolumn{5}{c}{Depth} &\multicolumn{1}{c}{MTL}\\
\cmidrule(lr){3-4} \cmidrule{5-9} \cmidrule(lr){10-10}
\multicolumn{2}{c}{}  &pAcc &mIoU   &\cellcolor{mycyan}AbR &\cellcolor{mycyan}RMSE &\cellcolor{mypink}{$\delta_1$}    &\cellcolor{mypink}$\delta_2$  &\cellcolor{mypink}$\delta_3$ 
&$\Delta M(\%)$\\
\midrule
\multirow{7}{*}{\rotatebox{90}{Cityscapes}}   
    &STL\_Seg       &95.7   &76.0   &-      &-              &-      &-      &-  &0.0\\ 
    &STL\_Depth     &-      &-      &0.694 &14.36 &46.9  &70.8  &81.5  &0.0\\
    &JTL \cite{kokkinos2017ubernet} &95.6  &75.5  &0.372 &8.646 &56.4  &81.1  &91.0 &+19.6\\
    &SemiSD \cite{thai2018adaptseg}  &95.6   &75.8   &0.349 &7.893 &59.3  &81.6  &91.5 &+22.3\\
    &SemiMTL (M1)                   &95.7   &76.2   &0.356 &7.959 &58.8  &81.4  &91.3 &+22.4\\
    &SemiMTL (M3)                   &95.7   &76.4   &0.341 &7.645 &59.5  &81.9  &91.7 &+23.7\\
    &SemiMTL            &\textbf{95.8}   &\textbf{76.9}   &\textbf{0.334} &\textbf{7.558}  &\textbf{61.4}  &\textbf{83.0}  &\textbf{91.9} &\textbf{+24.3}\\
\midrule
\multirow{7}{*}{\rotatebox{90}{Synscapes}}   
    &STL\_Seg       &90.9   &61.9   &-      &-      &-      &-      &- &0.0\\
    &STL\_Depth     &-      &-      &0.505  &6.214  &85.4  &94.1  &96.8  &0.0\\ 
    &JTL \cite{kokkinos2017ubernet} &91.3  &63.4   &0.486  &5.307 &87.4  &95.8  &96.8 &+8.5\\
    &SemiSD \cite{thai2018adaptseg}  &90.9  &62.8   &0.449 &5.183 &88.1  &95.8  &97.8 &+9.0\\
    &SemiMTL (M1)     &91.5   &\textbf{65.8}       &0.411 &5.153 &86.8  &94.1  &96.9 &+11.7\\
    &SemiMTL (M3)     &\textbf{91.6}   &65.5       &0.407 &5.157 &87.7  &95.2  &96.3 &+11.2\\
    &SemiMTL            &91.4  &65.4   &\textbf{0.380} &\textbf{5.056}  &\textbf{88.5}  &\textbf{95.9}  &\textbf{97.9}  &\textbf{+12.1}\\
\bottomrule
\end{tabular}
\end{table}

\subsection{Ablation Study}
\label{sec:exp_ablation}

To analyze the proposed approach thoroughly, we present the ablation study on the cross-domain setting in Table \ref{tab:across_domains}. We consider two baselines (JTL \cite{kokkinos2017ubernet} and SemiSD \cite{thai2018adaptseg, hung2018semiseg}) and different variants of the SemiMTL approach.
As stated in Section \ref{sec:semimtl_objective} and Figure \ref{fig:adversarial_modes}, the alternatives include (i) SemiMTL (M1): aligning the task output in unlabeled datasets to the prediction distribution in labeled datasets; (ii) SemiMTL (M2): encouraging the output to be similar to the ground-truth distribution, which is our default mode denoted as SemiMTL; (iii) SemiMTL (M3): enforcing the output to be close to the joint distribution of labeled prediction and ground-truth.

\noindent\textbf{Effect of Adversarial Training.} We first analyze the effect of directly utilizing the common adversarial learning method into the SemiMTL framework. 
We apply the adversarial scheme \cite{thai2018adaptseg, hung2018semiseg} to construct the discriminators of both tasks for semi-supervision, denoted as SemiSD, which only distinguishes the ground-truth and prediction distributions without considering the domain gap problem. 
Table \ref{tab:across_domains} shows that the SemiSD scheme performs better than the JTL baseline on both tasks of the Cityscapes dataset, but decreasing the segmentation IoU by 0.6\% on the Synscapes dataset.
These results show that a direct adversarial training lacks the generalization ability for tasks across domains.

\noindent\textbf{Effect of Domain-aware Module.} We further evaluate the effect of three different domain-aware modules which incorporate the domain information into training the discriminators. 
Table \ref{tab:across_domains} illustrates that these variants of the SemiMTL model all perform better than the JTL and SemiSD methods on the segmentation task of both datasets.
There is a slight performance loss in the SemiMTL (M1) model for the depth task, which shows that an ambiguous alignment to non-ground-truth distributions may not an effective way for the low-level tasks. 
However, the SemiMTL model performs consistently better than all baseline methods for all metrics on both datasets, indicating that the proposed semi-supervised MTL framework and domain-aware discriminators can learn more effective features and improve the performance of both tasks across domains.

\vspace{-2mm}
\section{Conclusions}
\label{sec:conclusions}
\vspace{-2mm}
In this paper, we present a new SemiMTL setting to address the multi-tasking across datasets. 
The proposed method is able to leverage the supervisory information from different domains and optimize all tasks simultaneously in a MTL model across datasets. 
We then introduce a domain-aware adversarial learning approach and various alignment modes to alleviate the domain discrepancy issue among datasets. 
We apply our SemiMTL model to two dense prediction tasks (semantic segmentation and depth estimation) on different challenging benchmarks.
Experimental results demonstrate the proposed SemiMTL method performs favorably against the state-of-the-art approaches.

{\small
\bibliographystyle{ieee_fullname}
\bibliography{egpaper}
}

\end{document}